\documentclass[conference]{IEEEtran}
\IEEEoverridecommandlockouts

\usepackage{cite}
\usepackage{amsmath,amssymb,amsfonts}
\usepackage{algorithmic}
\usepackage{graphicx}
\usepackage{textcomp}
\usepackage{xcolor}
\usepackage{float}
\usepackage{diagbox}
\usepackage{tikz}
\usepackage{pgfpages}
\usepackage{pgfplots}
\usepackage{booktabs}
\usepackage{comment}

\excludecomment{uniform-dev-model} % Hide all figures for the uniform deviation model

\begin{document}

%\title{Energy Gains of Systems with Faulty Memory for Modern Deep Neural Networks}
\title{Training Modern Deep Neural Networks for Memory-Fault Robustness}
%\thanks*{This work was supported by The Institute for Data Valorisation (IVADO).}

% NOTE: ISCAS has a double-blind review
\author{Ghouthi Boukli Hacene$^{1,2}$, Fran\c{c}ois Leduc-Primeau$^3$, Amal Ben Soussia$^2$, \\
        Vincent Gripon$^{1,2}$ and Fran\c{c}ois Gagnon$^4$,\\
        \footnotesize
        $^1$ Universit\'e de Montr\'eal, MILA 
        \hspace{5pt} $^2$ IMT Atlantique, Lab-STICC\\
        \footnotesize
        $^3$ Dept. of Electrical Engineering, \'Ecole Polytechnique de Montr\'eal  
        \hspace{5pt}
        $^4$ Dept. of Electrical Engineering, \'Ecole de technologie sup\'erieure de Montr\'eal%\\
        \thanks{This work was supported by The Institute for Data Valorisation (IVADO).}
\vspace{-5pt}
        }
%\thanks{This work was supported by The Institute for Data Valorisation (IVADO).}
\maketitle

\begin{abstract}
Because deep neural networks (DNNs) rely on a large number of parameters and computations, their implementation in energy-constrained systems is challenging. In this paper, we investigate the solution of reducing the supply voltage of the memories used in the system, which results in bit-cell faults. 
We explore the robustness of state-of-the-art DNN architectures towards such defects and propose a regularizer meant to mitigate their effects on accuracy. Our experiments clearly demonstrate the interest of operating the system in a faulty regime to save energy without reducing accuracy.
\end{abstract}

%\begin{IEEEkeywords}
%Deep neural networks, embedded systems, voltage scaling, implementation errors.\end{IEEEkeywords}

\section{Introduction}

Deep Neural Networks~\cite{lecun2015deep} (DNNs) are the golden standard for many challenges in machine learning. Thanks to the large number of trainable parameters that they provide, DNNs can capture the complexity of large training datasets and generalize to previously unseen examples.

Among the many applications for DNNs, many are in the field of embedded systems. Examples include monitoring of health signals, human-machine interfaces, autonomous drones, and smartphone applications. 
Many such embedded applications cannot rely on cloud-based processing because of stringent latency constraints or privacy issues. Even when cloud processing is an option, processing in-device or at the network edge can be useful to save network bandwidth.
The energy consumption of the inference task is thus a major concern.
Unfortunately, because state-of-the-art DNN architectures are composed of a large number of trained parameters, the inference step typically requires significant energy to achieve accurate results on challenging tasks, with a large part of the energy complexity being associated with the memory accesses required to retrieve the parameters and save temporary results.

Since off-chip memory accesses consume significant energy, a first step for reducing energy consumption consists in storing all parameters and temporary results on the same chip as the hardware accelerator, using static random access memories (SRAMs), or potentially embedded dynamic RAMs~\cite{chen:2014}.
However, even in this case, the energy consumed by memory accesses still represents 30--60\% of the total energy~\cite{kim:2018}.
An effective way of lowering energy consumption of both memory and logic circuits is to reduce the supply voltage, but this has the effect of increasing the sensitivity of the circuits to fabrication variations, causing bit-cell failures in SRAMs. When approaching the minimum energy operating point of SRAMs, the failure rates increase by several orders of magnitude compared to operating at the nominal supply~\cite{dreslinski:2010}.
However, even such large bit-cell failure rates are not necessarily catastrophic if appropriate mechanisms are in place to safeguard the operation of the system.

DNNs naturally exhibit a limited amount of fault tolerance, as noted for instance in \cite{vialatte:2017,jiao:2017}, and there is a growing body of work that studies the operation of DNN inference hardware built using faulty memories. We review several contributions in Section~\ref{sec:relatedwork}.
The aim of this paper is to investigate the ability to decrease the energy consumption of DNN accelerators by allowing the memories used for storing weights and activations to operate in a faulty regime, thus introducing \emph{deviations} on the stored values.
We rely on simple but realistic energy-deviation models to explore the impact of memory failures on classification accuracy, and ultimately on energy consumption.

We quantify the impact on robustness of several design aspects of state-of-the-art deep architectures in order to identify whether these aspects should be targeted when designing robust architectures. Specifically, we consider the choice of general architecture, how the depth of a layer impacts its robustness, and the impact of faults occurring in the storage of weights or of neuron activations. 
Interestingly, we find that different architectures provide varying degrees of robustness.

We then consider whether faulty operation can lead to a reduction in power consumption.
Importantly, we compare the energy consumption with a reliable reference implementation that achieves the same application performance. We show that using a faulty implementation to reduce energy consumption at the cost of a reduction in accuracy is not necessarily beneficial, even when the loss in accuracy appears small. Indeed, for state-of-the-art architectures, accepting even a 1\% reduction in accuracy can significantly reduce the number of parameters required by a reliable implementation.
It it thus essential to evaluate the improvement provided by a faulty architecture at the same accuracy. 
Nonetheless, we show that faulty operation can reduce energy consumption when the fault statistics are taken into account during training.

The outline of the paper is as follows. 
Section~\ref{sec:relatedwork} briefly reviews related work. Section~\ref{sec:model} introduces the deviation models, which represent the impact of circuit faults on the algorithm. Section~\ref{sec:experiments} presents an exploration of the design space for faulty-memory implementations of modern DNNs. Section~\ref{sec:strategies} proposes a regularizer to increase the robustness of DNNs to deviations. Section~\ref{sec:conclusion} provides some conclusions.

\section{Related Work}\label{sec:relatedwork}
The idea of exploiting fault tolerance to improve the energy efficiency of neural networks has attracted a significant number of contributions. An early investigation of the effect of transistor-level defects on neural networks was performed in \cite{temam:2012}.
More recently, circuit-level methods for improving the application performance of faulty implementations have been proposed. One approach consists in using \emph{razor} flip-flops to detect faults and selectively apply a compensation mechanism. When memory faults can be detected at the bit level, a \emph{bit masking} technique can be applied to ensure that errors always reduce the magnitude of weights, helping to decrease the impact of the errors on performance~\cite{reagen:2016,whatmough:2017}. Similarly, razor flip-flops can be used to compensate timing violations occurring in the datapath by dropping the next operation, which effectively sets its weight parameter to zero~\cite{zhang:2018}.
Finally, a low-precision replica can be added to computations units to bound the maximum error that can be introduced by a faulty processing unit~\cite{lin:2016}.

To the best of our knowledge, few papers investigate the effect of training deep architectures to increase fault robustness. One notable exception is \cite{kim:2018}, which proposes modifying the training procedure to take into account bit flips occurring in SRAMs, and present results on the MNIST benchmark~\cite{lecun1998mnist}.
The effect of faults occurring in the storage of the input is also considered in \cite{yang:2017}, and \cite{wang:2015} proposes on-chip learning for support-vector machines, while decreasing the learning effort using \emph{active learning}.
Finally, a slightly different problem is considered in \cite{liu:2017,xia:2018}, where the network is trained to compensate for known defect locations.

Another line of work consists in compressing models to reduce memory usage and number of computations. There are mainly three ways to achieve this. A first one is to quantize weights, using in the extreme case only one bit per weight and per activation~\cite{courbariaux2015binaryconnect, courbariaux2016binarized, soulie2016compression}. While the process has proven very efficient on old and somewhat redundant architectures, it can drastically affect accuracy when performed on already compressed architectures. A second way to compress DNNs is to prune the weights, significantly reducing the number of parameters to be stored~\cite{han2015learning}. A last line of work consists in factorizing weights, so that they can be used to perform multiple computations throughout the processing of an input~\cite{han2015deep, kim2015compression, hou2016loss}. However, in modern architectures the number of weights is only a small portion of the memory, as activations of neurons can be as many and even more if the batch size is large, that is if several inputs are processed in parallel.

\section{Energy-Deviation Model}
\label{sec:model}

We focus on the energy consumed by memory accesses, and assume that the amount of energy required to perform an inference task is proportional to the number of accesses. We thus define a base energy metric $E_o$ that is the sum of the number of parameters and of the number of activation values generated during the inference.

To decrease the energy consumption of on-chip memories, we consider reducing the supply voltage, which in turn causes some bit cells to fail. 
%A typical model for such bit-cell failures is the stuck-at model, for which a cell is either stuck at a logical 0 or at a logical 1 value. Assuming that the two stuck-at outcomes are equiprobable and that stored bits are also equiprobable, we can simplify the model to a binary symmetric channel model where stored bits have a probability $p$ of being flipped.
%
In order to investigate the general behavior of DNNs implemented with faulty memory, we need a model linking the bit-cell fault probability $p$ and the energy consumed by memory accesses. We denote by $0 \leq \eta \leq 1$ the normalized energy consumption of the memory, where the normalization is with respect to the energy consumption of the reliable memory (such that the energy is given by $\eta E_o$).
Note that we can obtain a simple upper bound for $p$ from the fact that instead of using a faulty memory, we could store only a fraction $\eta$ of the data while declaring the missing bit-cells as faulty, which yields a linearly decreasing $p(\eta)$. % Note: Specifically $p(\eta)=(1-\eta)$

Based on reliability data published in \cite[Fig.7]{dreslinski:2010}, we will assume that the energy-reliability function takes the exponential form
\begin{equation}\label{eq:energymodel}
p(\eta) = e^{-a\eta} \, .
\end{equation}
In order to obtain a specific value of parameter $a$ for illustrative purposes, we select $a$ to fit the energy data reported in \cite[Fig.1]{chen:2010} and the reliability from \cite[Fig.7]{dreslinski:2010} for 65nm CMOS SRAM cells at $V_\mathrm{DD} \in \{0.5, 1.1\}$. Performing the fit by minimizing the sum of the relative squared error yields $a=12.8$.
Specific energy gains will vary based on the value of $a$, but in this paper we are only interested in identifying general trends.

The manner in which memory faults introduce deviations during inference depends on the strategy being used to cope with faults. We consider the case where bit-cell faults can be detected, and use the bit masking (BM) approach proposed in \cite{reagen:2016}. When a fault is detected on the sign bit of a value, this value is replaced with zero. In the case of failures on any other bits, the affected bit values are replaced with the sign bit, causing the value to deviate towards zero. We consider that all bit cells have an equal fault probability $p$.
When using the deviation model in simulations, we assume that values are quantized on 8 bits. However, for a fair comparison with the reliable implementations that use a floating-point representation, we compute the deviation in the quantized domain, but apply it on the floating-point representation. 
%, that is, we generate a real-valued $x'$ that is the faulty version of $x$ as $x'=x+(\tilde{y}-\tilde{x})$, where $\tilde{x}$ is the quantized version of $x$, and $\tilde{y}$ is the quantized value after applying deviations.
Unless otherwise mentioned, we consider that faults affect both the weights and the neuron activations. Note that activations are known to be positive since they are generated by a ReLU function. Therefore, we assume that their sign bit cannot be affected.

In this work, we use our deviation model during the training phase to increase the robustness of networks and thus their energy efficiency.
Because training is computationally intensive, we propose to simplify the BM deviation model used during training to speed up the process.
Since the BM approach always causes values to deviate towards zero, we propose to approximate it using a deviation model that will be referred to as the \emph{erasure} model, for which each value has a probability $p_e$ of being set to zero. We then need to choose $p_e$ to best approximate the effect of the BM model.
We can first note that in the case of weight parameters, the BM model sets the faulty value to zero in case of a sign-bit fault, which occurs with probability $p$. Therefore, we clearly need $p_e > p$. During training, this process is similar to dropout~\cite{srivastava2014dropout}, but it is used to increase the robustness of networks, and not to prevent overfitting. To find the best choice of $p_e$ to approximate the BM model, we evaluate the performance of both models on the test set and choose the value of $p_e$ that best predicts the accuracy of the network under the BM model.

\section{Design-Space Exploration for Faulty Implementations}
\label{sec:experiments}
% This section investigates the effect of the following aspects on robustness:
% 1) effect of the general architecture
% 2) the layer (or blocks of layers) in which the faults occur
% 3) whether only weights, only activations, or both weights and activations are affected by faults.

\subsection{Choice of architecture and dataset}

We perform experiments using the CIFAR10 dataset~\cite{krizhevsky2014cifar} made of tiny color images of 32$\times$32 pixels. We first compare four architectures, namely PreActResNet18~\cite{he2016identity}, MobileNetV2~\cite{sandler2018mobilenetv2}, SENet18~\cite{hu2017squeeze} and ResNet18~\cite{he2016deep}, which are all modern architectures achieving good accuracy on CIFAR10. Table~\ref{tbl:memaccess} shows for each architecture the number of weights (parameters) and activation values of neurons that must be retrieved from memory for processing one input, and the accuracy achieved by that architecture.

\begin{table}
    \caption{Number of memory accesses and accuracy by architecture}\label{tbl:memaccess}
    \begin{center}
        \vspace{-10pt}
        \begin{tabular}{cccc}
        \toprule
        Architecture & Parameters & Activations & Accuracy\\
        \midrule
        PreActResNet18 \cite{he2016identity} & $11.2 \times 10^6$ & $0.55 \times 10^6$ &   94.87\%\\ % activations: 548864
        MobileNetV2 \cite{sandler2018mobilenetv2} & $2.30 \times 10^6$ & $1.53 \times 10^6$ & 93.80\% \\ % parameters: $2296922$, activations: 1534976
        SENet18 \cite{hu2017squeeze} & $11.3 \times 10^6$ & $0.86 \times 10^6$ & 94.77\%\\ % parameters: $11260354$, activations: 860160
        ResNet18 \cite{he2016deep} & $11.2 \times 10^6$ & $0.56 \times 10^6$ & 94.86\%\\ % parameters: $11173962$, activations: 557056
        \bottomrule
        \end{tabular}
    \end{center}
\end{table}

\begin{figure}[t]
  \begin{center}
    \begin{tikzpicture}
       \begin{scope}[scale=1, yshift=6cm]
        \begin{axis}[
            xlabel=$p$,
            xmode=log,
            ylabel=Test set accuracy (\%),
            ymin=60,
            legend pos=south west,
            legend style={font=\footnotesize},
            legend cell align={left}, % align legend text on the left
            width=8cm,height=6cm
            ]
          \addlegendentry{PreActResNet18}
          \addplot coordinates
          {(0.001,94.762)(0.002,94.657)(0.005,94.474)(0.01,93.905)(0.02,92.883)(0.05,82.846)
          %(0.1,45.428)(0.2,13.231)(0.5,10.0)
          };
          \addlegendentry{MobileNetV2}
          \addplot coordinates {(0.001,92.54100000000001)(0.002,91.681)(0.005,89.548)(0.01,85.461)(0.02,59.14899999999999)(0.05,19.375)%(0.1,10.425)(0.2,9.992)(0.5,9.976)
          };
          \addlegendentry{SENet18}
          \addplot coordinates {(0.001,94.77000000000001)(0.002,94.36600000000001)(0.005,93.393)(0.01,90.857)(0.02,82.5)(0.05,43.252)%(0.1,15.266)(0.2,10.141)(0.5,9.997)
          };
          \addlegendentry{ResNet18}
          \addplot coordinates {(0.001,94.86300000000001)(0.002,94.747)(0.005,94.559)(0.01,93.908)(0.02,91.91)(0.05,70.09400000000001)%(0.1,18.69)(0.2,11.157)(0.5,9.91)
          };
          %% \addplot table {resgain.txt};
          %% \addlegendentry{Adding shortcuts}
        \end{axis}
      \end{scope}
    \end{tikzpicture}
  \end{center}
  
  \vspace{-0.45cm}
  \caption{Impact of the architecture on the robustness under BM deviations.}
  \label{fig:arch_erasure_p}
\end{figure}
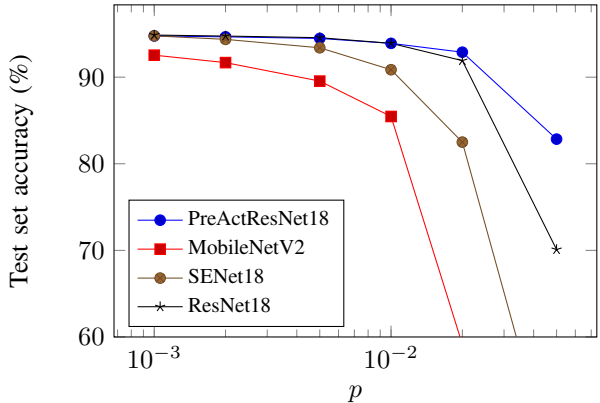  

In Fig.~\ref{fig:arch_erasure_p}, we compare the robustness of the above-mentioned architectures when the parameters and activations are affected by the BM deviation model. We observe that some architectures are inherently more robust than others, and that this does not depend solely on the global number of parameters.
In Fig.~\ref{fig:arch_erasure_energy}, we plot the accuracy in terms of the energy $\eta E_o$ per inference, where the base energy $E_o$ corresponds to the sum of the parameter and activation columns of Table~\ref{tbl:memaccess}, and the fault probability is obtained from the normalized energy $\eta$ using \eqref{eq:energymodel}.
We observe that PreActResNet18 provides a very interesting trade-off between accuracy, memory accesses and robustness to BM. Therefore we choose to focus on this architecture for the remaining experiments.

\def\xref{11720010}
\begin{figure}[t]
  \begin{center}
    \begin{tikzpicture}
       \begin{scope}[scale=1, yshift=6cm]
        \begin{axis}[
            xlabel=Normalized energy,
            %xmode=log,
            ymin=90,
            ylabel=Test set accuracy (\%),
            legend pos=south east,
            legend style={font=\footnotesize},
            legend cell align={left}, % align legend text on the left
            width=8cm,height=6cm
            ]
          \addlegendentry{PreActResNet18}
          \addplot coordinates
          {(11750000/\xref,94.87)(5704816/\xref,94.657)(4863689/\xref,94.474)(4227402/\xref,93.905)(3591114/\xref,92.883)(2749988/\xref,82.846)
          %(0.1,45.428)(0.2,13.231)(0.5,10.0)
          };
          \addlegendentry{MobileNetV2}
          \addplot coordinates {(3831898/\xref,93.8)(2066929/\xref,92.5)(1860448/\xref,91.681)(1586141/\xref,89.548)(1378635/\xref,85.461)(1171130/\xref,59.14899999999999)(896822/\xref,19.375)
          %(0.1,10.425)(0.2,9.992)(0.5,9.976)
          };
          \addlegendentry{SENet18}
          \addplot coordinates {(12160000/\xref,94.770)(5903877/\xref,94.3660)(5033401/\xref,93.393)(4374911/\xref,90.857)(3716421/\xref,82.5)(2845945/\xref,43.252)
          %(0.1,15.266)(0.2,10.141)(0.5,9.997)
          };
          \addlegendentry{ResNet18}
          \addplot coordinates {(11750000/\xref,94.86300000000001)(5704816/\xref,94.747)(4863689/\xref,94.559)(4227402/\xref,93.908)(3591114/\xref,91.91)(2749988/\xref,70.09400000000001)%(0.1,18.69)(0.2,11.157)(0.5,9.91)
          };
          %% \addplot table {resgain.txt};
          %% \addlegendentry{Adding shortcuts}
        \end{axis}
      \end{scope}
    \end{tikzpicture}
  \end{center}
  
  \vspace{-0.45cm}
  \caption{Energy consumption of different architectures under BM deviations.}
  %\francois{TODO: ajouter des points sur la courbe Preact-Resnet}}
  \label{fig:arch_erasure_energy}
\end{figure}
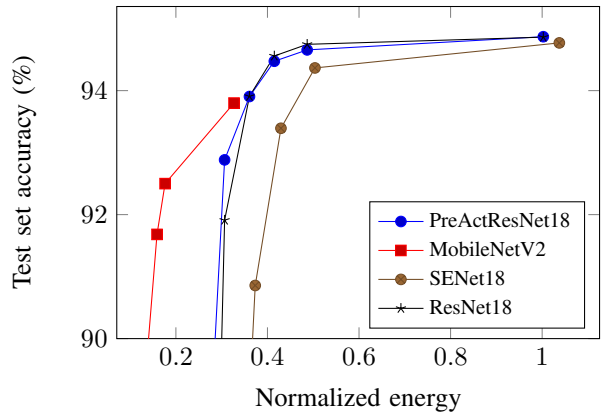

\subsection{Comparison of the BM and erasure models}\label{subsec:BMvsErasures}

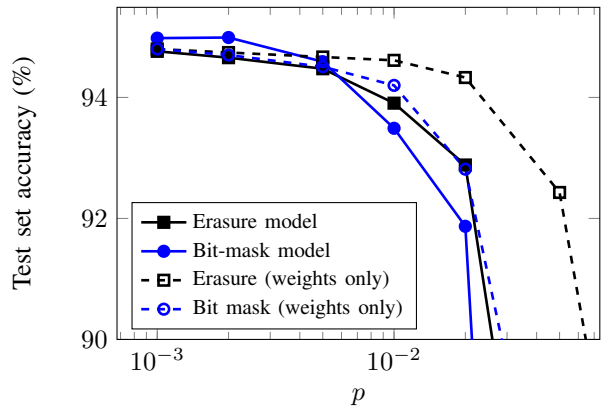
\begin{figure}[t]
  \begin{center}
    \begin{tikzpicture}
           \begin{scope}[scale=1, yshift=6cm]
        \begin{axis}[
            xlabel=$p$,
            ylabel=Test set accuracy (\%),
            ymin=90,
            legend style={font=\footnotesize},
            legend cell align={left}, % align legend text on the left
            legend pos = south west,
            xmode = log,
            width=8cm,height=6cm]
          \addlegendentry{Erasure model}
          \addplot[black,
                   mark=square*,
                   line width = 1pt] coordinates
          {(0.001,94.762)(0.002,94.657)(0.005,94.474)(0.01,93.905)(0.02,92.883)(0.05,82.846)
          (0.1,45.428)(0.2,13.231)(0.5,10.0)
          };
          
          \addlegendentry{Bit-mask model}
          \addplot[blue,
                 mark=*,
                 line width = 1pt] coordinates 
          {(0.001,94.98)(0.002,94.99)(0.005,94.59)(0.01,93.49)(0.02,91.87)(0.05,65.3)(0.1,17.0)(0.2,10.0)(0.5,10.0)};
        
          \addlegendentry{Erasure (weights only)}
          \addplot[black,
                   style=dashed,
                   line width = 1pt,
                   mark=square, mark options={solid}] coordinates
          {(0.001,94.8)(0.002,94.74199999999999)(0.005,94.66799999999999)(0.01,94.61299999999999)(0.02,94.329)(0.05,92.429)(0.1,85.82700000000001)(0.2,38.059999999999995)(0.5,10.283000000000001)
          };
          
         \addlegendentry{Bit mask (weights only)}
          \addplot[blue,
                 style=dashed,
                 line width = 1pt,
                 mark=o, mark options={solid}] coordinates {(0.001,94.8)(0.002,94.7)(0.005,94.507)(0.01,94.19900000000001)(0.02,92.81400000000001)(0.05,85.53299999999999)(0.1,37.559000000000005)(0.2,10.221)(0.5,10.0)
                 };
        \end{axis}
      \end{scope}
    \end{tikzpicture}
  \end{center}

\vspace{-0.45cm}
  \caption{Impact of memory faults on accuracy for different deviation models.
  %\francois{TODO: Accuracy improves when p goes from 0.001 to 0.002 for weight-only erasures!}
  }
  \label{fig:robustness_devmodel}
\end{figure}

As motivated in Section~\ref{sec:model}, we are interested in comparing the effects of BM and erasures on the chosen architecture. Results are depicted in Fig.~\ref{fig:robustness_devmodel}. 
Since the BM model affects weights and activations differently and since PreActResNet18 has about $20\times$ more weights than activation values, we focus on matching the accuracy of the two models when only weights are affected by deviations. We observe for this case that the BM and erasure models have a similar effect, provided that $p_e = 2 p$, suggesting that using erasures as a proxy to model the deviations induced by BM is a reasonable option. This relation will be used in Section~\ref{sec:strategies} to train networks to be more resilient to BM deviations.

\subsection{Relative importance of layer depth}\label{subsec:layerdepth}

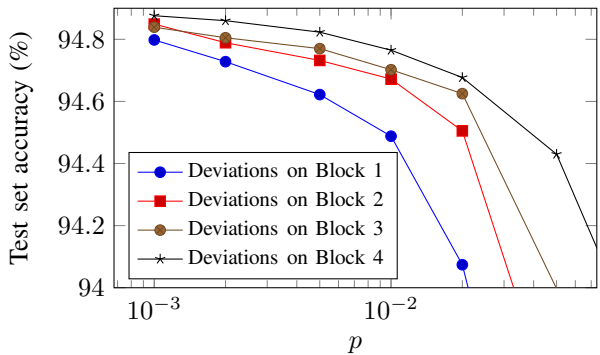
\begin{figure}[t]
  \begin{center}
    \begin{tikzpicture}
           \begin{scope}[scale=1, yshift=6cm]
        \begin{axis}[
            xlabel=$p$,
            xmode=log,
            ymin=94,
            ymax=94.9,
            ylabel=Test set accuracy (\%),
            legend pos=south west,
            legend style={font=\footnotesize},
            legend cell align={left}, % align legend text on the left
            width=8cm,height=5.3cm
            ]
          \addlegendentry{Deviations on Block 1}
          \addplot coordinates {(0.001,94.798)(0.002,94.728)(0.005,94.62199999999999)(0.01,94.48799999999999)(0.02,94.074)(0.05,92.87199999999999)(0.1,89.523)(0.2,79.20100000000001)
          %(0.5,35.261)
          }; 
          \addlegendentry{Deviations on Block 2}
          \addplot coordinates {(0.001,94.849)(0.002,94.789)(0.005,94.732)(0.01,94.672)(0.02,94.505)(0.05,93.578)(0.1,88.23400000000001)(0.2,49.529999999999994)
          %(0.5,10.168000000000001)
          };  
          \addlegendentry{Deviations on Block 3}
          \addplot coordinates {(0.001,94.839)(0.002,94.80499999999999)(0.005,94.77000000000001)(0.01,94.70199999999998)(0.02,94.625)(0.05,93.99700000000003)(0.1,92.251)(0.2,82.788)
          %(0.5,21.797)
          };
          \addlegendentry{Deviations on Block 4}
          \addplot coordinates {(0.001,94.876)(0.002,94.86000000000001)(0.005,94.82300000000001)(0.01,94.76500000000001)(0.02,94.67699999999999)(0.05,94.43000000000002)(0.1,93.896)(0.2,91.04899999999999)
          %(0.5,16.0)
          };
          %% \addplot table {resgain.txt};
          %% \addlegendentry{Adding shortcuts}
        \end{axis}
      \end{scope}
    \end{tikzpicture}
  \end{center}
  
  \vspace{-0.45cm}
  \caption{Impact on accuracy of BM deviations applied to different stages of the network, ``Block 1'' being the first and ``Block 4'' the last.}
  \label{fig:robustness_stages}
  \vspace{-.2cm}
\end{figure}

In a new series of experiments, we aim at identifying the relative robustness of various parts of the architecture under BM deviations. To this end, we introduce deviations on only a portion of the network. Since PreActResNet18 is composed of 4 sequential blocks (made of convolutional layers and shortcuts), we apply BM deviations to the weights and activations of only one block at a time.
Results are depicted in Fig.~\ref{fig:robustness_stages}. We observe that all parts of the network are sensitive to deviations.
Interestingly, in the region of small accuracy degradation shown in Fig.~\ref{fig:robustness_stages}, robustness increases monotonically with the depth of the block.
%This figure advocates that it would be beneficial to assign varying supply voltages to the memories associated with each block, so that the fault likelihood is allowed to increase with the depth of the block.
%
We thus consider exploiting the varying robustness of the layers to improve energy consumption by assigning different operating points to each block. Denoting by $p_{\mathrm{B}i}$ the fault probability assigned to block $i$, we note from Fig.~\ref{fig:robustness_stages} that at a high accuracy of 94.8\%, 
$p_\mathrm{B4} =\allowbreak 5 p_\mathrm{B3} =\allowbreak 5 p_\mathrm{B2} =\allowbreak 10 p_\mathrm{B1}$. The number of parameters associated with each block, which in order of block is 
$[1.5, 5.2, 21, 84] \times 10^5$, %$[147456, 524288, 2097152, 8388608]$
also varies over a wide range. Following intuition, blocks that are more robust also have more parameters. As shown in Fig.~\ref{fig:energy_strategies} (curves labeled ``Diff. Fault.''), applying this fault-rate policy significantly improves the energy efficiency of the standard network.

%\francois{We should extract the number of parameters per block, and discuss whether further energy gains are possible if the "p" is chosen independently for the parameters of each block. A simple way to use the information in fig.~\ref{fig:robustness_stages} is as follows: pick some accuracy target, extract the relative fault tolerance of each block, then impose that relative p constraint so that the deviation parameter remains one-dimensional. So for instance with a target of 94.8\%, we would get $p_\mathrm{B4}=5 p_\mathrm{B2}=5 p_\mathrm{B3}=10 p_\mathrm{B1}$. (I focused the plot on accuracy values above 94\% to illustrate this relationship.) Then use this to generate an accuracy-energy curve. We should be able to get additional energy gains this way...}\vincent{I do not know if it is feasible, it would require to draw the same curve but this time using the regularizer to be exploitable in Fig 5. Ghouthi, what do you think?}

\begin{uniform-dev-model}
\begin{figure}[t]
  \begin{center}
    \begin{tikzpicture}
           \begin{scope}[scale=1, yshift=6cm]
        \begin{axis}[
            xlabel=$p$,
            xmode=log,
            ylabel=Test set accuracy (\%),
            legend pos=south west,
            ]
          \addlegendentry{Error noise on block1}
          \addplot coordinates {(0.001,94.757)(0.002,94.74100000000001)(0.005,94.65099999999998)(0.01,94.56099999999999)(0.02,94.126)(0.05,92.911)(0.1,90.07300000000001)(0.2,76.86599999999999)(0.5,29.275)}; 
          \addlegendentry{Error noise on block2}
          \addplot coordinates {(0.001,94.807)(0.002,94.776)(0.005,94.718)(0.01,94.643)(0.02,94.48599999999999)(0.05,92.84700000000001)(0.1,84.07000000000001)(0.2,36.361000000000004)(0.5,10.044)};  
          \addlegendentry{Error noise on block3}
          \addplot coordinates {(0.001,94.83599999999998)(0.002,94.776)(0.005,94.733)(0.01,94.648)(0.02,94.542)(0.05,93.70500000000001)(0.1,90.705)(0.2,65.66400000000002)(0.5,12.221)};
          \addlegendentry{Error noise on block4}
          \addplot coordinates {(0.001,94.86500000000001)(0.002,94.84100000000002)(0.005,94.81700000000001)(0.01,94.75200000000001)(0.02,94.69800000000001)(0.05,94.47099999999999)(0.1,94.044)(0.2,91.838)(0.5,25.349)};
          %% \addplot table {resgain.txt};
          %% \addlegendentry{Adding shortcuts}
        \end{axis}
      \end{scope}
    \end{tikzpicture}
  \end{center}
  \caption{Impact on accuracy of uniform-output deviations applied to different stages of the network.}
  \label{fig:layerdepth-error}
\end{figure}
\end{uniform-dev-model}

\subsection{Impact of the number of parameters in the architecture}
The number of parameters can be easily adapted by modifying the number of feature maps in the convolutional layers. If the number of feature maps of each convolutional layer is multiplied by $k$, then the total number of parameters will be roughly multiplied by $k^2$, as the number of parameters in a convolutional layer increases linearly with both the number of input feature maps and the number of output feature maps.

We train two variants of the PreActResNet architecture in which the original number $F$ of feature maps is multiplied by $1/2$ and $1/\sqrt{2}$.
These networks are used to provide a reference for the performance achieved with faulty implementations. The $F/2$ and $F/\sqrt{2}$ networks achieve respectively an accuracy of 93.45\% and 94.41\% under reliable implementations, illustrating the fact that significant energy reductions can be obtained easily if a reduced accuracy is acceptable.

\section{Proposed Regularizer}
\label{sec:strategies}

All previous experiments confirm that modern DNN architectures can tolerate some amount of deviations. However, in all the scenarios considered, we observe a sharp drop in performance as soon as the probability $p$ of defect becomes too large or the energy too small. To improve the robustness to deviations, we consider training the networks in the same conditions they are used in, which means that we apply erasures during the forward pass of the training phase. We call this method the erasure regularizer. Note that the reason that we use erasure rather than BM deviations is to speed up the training process.

In Fig.~\ref{fig:energy_strategies}, we plot the accuracy of the networks as a function of the energy they use. We compare reliable implementations of networks with varying number of parameters with the performance obtained when reducing the supply voltage of memories. 
For the specific energy model discussed in Section~\ref{sec:model}, the best energy reduction obtained by the faulty implementations with $F$ feature maps is $1.5\times$ for the network with standard training, achieved at an accuracy of 94.76\% and a fault rate of $p=0.001$, while the best energy reduction obtained using the erasure regularizer is $2.3\times$ at an accuracy of 94.8\% and $p=0.01$.
Furthermore, additional gains can be obtained by combining the erasure regularizer with blockwise reliability assignment of Sect.~\ref{subsec:layerdepth}.
We thus see that training the network for robustness using the erasure regularizer can significantly improve the energy reduction obtained from faulty operation under the bit-masking model, at equal accuracy. 
As discussed in Sect.~\ref{subsec:BMvsErasures}, it is important to perform the training with the appropriate $p_e$ parameter: using an erasure regularizer with $p_e=p$ did not yield an improvement in robustness.

%This plot shows that for a given energy budget, significant energy savings can be obtained
%There are two interesting conclusions we can derive from this plot:
%\begin{enumerate}
%    \item For a given energy budget, it can be more efficient to reduce the supply voltage than considering lowering the number of parameters of the architecture, though it depends on the amount of energy that needs to be saved,
%    \item Using the erasure regularizer has a significant impact on robustness of the architecture, so that it allows reducing energy by a large margin while reaching the same accuracy.
%\end{enumerate}

\def\xref{11720010}
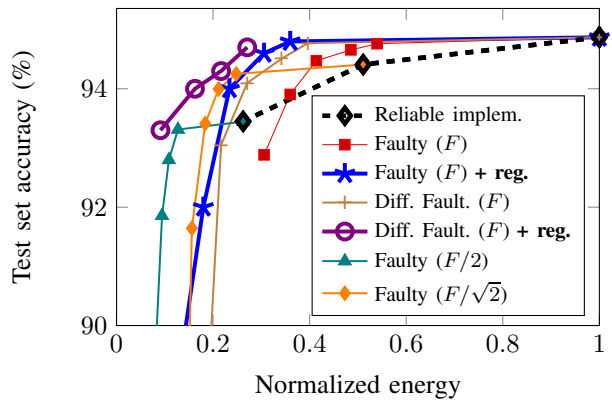
\begin{figure}[t]
  \begin{center}
    \begin{tikzpicture}
           \begin{scope}[scale=1, yshift=6cm]
        \begin{axis}[
            xlabel=Normalized energy,
            %xmode=log,
            ylabel=Test set accuracy (\%),
            ymin=90,
            xmax=1,
            legend pos=south east,
            legend style={font=\footnotesize},
            legend cell align={left}, % align legend text on the left
            width=8cm,height=5.8cm
            ]
          \addlegendentry{Reliable implem.}
          \addplot[black,
                   style=dashed,
                   mark=diamond,
                   mark size=3pt, mark options={solid},
                   line width = 1.75pt]
          coordinates {(11720010/\xref,94.87)(5988908/\xref,94.41)(3074999/\xref,93.45)};
          % interpolation of energy @accuracy=x (x>94.41):
          % (11720010-5493754)/(95-94.41)*(x-94.41)+5493754
          % @accuracy=94.8: 9.6094e+06
          % @accuracy=94.762: 9.2084e+06
          % @accuracy=94.657: 8.1003e+06
          % @accuracy=94.6: 7.4988e+06
          % @accuracy=94.474: 6.1691e+06
          
          \addlegendentry{Faulty ($F$)}
          \addplot coordinates {
          (11720010/\xref,94.87)
          (6324918/\xref,94.762) % p=???
          (5690254/\xref,94.657) % p=???
          (4851275/\xref,94.474) % p=???
          (4216612/\xref,93.905)
          (3581948/\xref,92.883)};
          % Energy gain @accuracy=94.762: 9.2084e+06/6324918 = 1.5x
          % Energy gain @accuracy=94.657: 8.1003e+06/5690254 = 1.4x
          % Energy gain @accuracy=94.474: 6.1691e+06/4851275 = 1.3x
          
          \addlegendentry{Faulty ($F$) {\bf + reg.}}
          \addplot[blue,
                   mark=star,
                   mark size=4pt,
                   line width = 1.5pt]
          coordinates {(11720010/\xref,94.87)
           (4216612/\xref,94.8) % p=0.01
           (3581948/\xref,94.6) % p=0.02
           (2742969/\xref,94)
           (2108306/\xref,92)
           (1473642/\xref,89)};
          % Energy gain @accuracy=94.8: 9.6094e+06/4216612 = 2.3x
          % Energy gain @accuracy=94.6: 7.4988e+06/3581948 = 2.1x

          \addlegendentry{Diff.~Fault.~($F$)}
          \addplot[brown,
                   mark=text, text mark=+,
                   mark size=3pt,
                   line width = 0.75pt]
          coordinates {(11720010/\xref,94.87)(4645901/\xref,94.77)(4011976/\xref,94.51)(3173973/\xref,94.1)(2540049/\xref,93.05)(1906124/\xref,84.83)};
          
          \addlegendentry{Diff.~Fault.~($F$) {\bf + reg.} }
          \addplot[violet,
                   mark=o,
                   mark size=3pt,
                   line width = 1.5pt]
          coordinates {(3173973/\xref,94.7)(2540049/\xref,94.3)(1906124/\xref,94)(1068121/\xref,93.3)};
          
          \addlegendentry{Faulty ($F/2$)} % i.e. n/4
          \addplot[teal,
                   mark=triangle*,
                   mark size=2.5pt,
                   line width = 0.75pt]
          coordinates {(3074999/\xref,93.44500000000001)(1492962/\xref,93.312)(1272837/\xref,92.799)(1106319/\xref,91.856)(939802/\xref,89.40299999999999)(719677/\xref,69.16499999999999)
          };
          
          \addlegendentry{Faulty ($F/\sqrt{2}$)} % i.e. n/2
          \addplot[orange,
                   mark=diamond*,
                   mark size=2.5pt,
                   line width = 0.75pt]
          coordinates {(5988908/\xref,94.41)(2907712/\xref,94.25199999999998)(2478994/\xref,93.99600000000001)(2154682/\xref,93.41199999999999)(1830370/\xref,91.64299999999999)(1401653/\xref,76.85)
          };
          
          %\addlegendentry{Faulty ($\sqrt{2}F$)} % i.e. 2n
          %\addplot[magenta,
          %         mark=*,
          %         mark size=2pt,
          %         line width = 0.75pt]
          %coordinates %{(23177817/\xref,94.684)(11253206/\xref,94.619)(9594017/\xref,94.43800000000002)(8338889/\xref,94.00399999999999)(7083761/\xref,92.397)(5424573/\xref,79.02900000000001)
          %};
          \addplot[black,
                   style=dashed,
                   mark=diamond,
                   mark size=3pt, mark options={solid},
                   line width = 1.75pt]
          coordinates {(11720010/\xref,94.87)(5988908/\xref,94.41)(3074999/\xref,93.45)};
          %\addlegendentry{Faulty ($2F$)}
          %\addplot coordinates {(45900000,94.968)(22285196,94.84899999999999)(18999434,94.675)(16513852,94.34299999999999)(14028269,93.104)(10742508,82.349)
          
          %};
          
          %% \addplot table {resgain.txt};
          %% \addlegendentry{Adding shortcuts}
        \end{axis}
      \end{scope}
    \end{tikzpicture}
  \end{center}
  
  \vspace{-.45cm}
  \caption{Energy consumption of the Preact-Resnet18 architecture under BM deviations. Each faulty implementation curve corresponds to a fixed network size, with the number of feature maps shown within parentheses.
  %\francois{TODO: re-run "faulty+reg" with more trials (and could we add a point between the 2nd and the 3rd?)}
  }
  \label{fig:energy_strategies}
  \vspace{-.2cm}
\end{figure}

\vspace{-.1cm}
\section{Conclusion}
\label{sec:conclusion}

In this work, we explored the possibility of exploiting the fault tolerance of deep neural networks to reduce the energy consumption of on-chip memories.
We showed that in some conditions, reducing the supply voltage can result in better accuracy for the same energy consumption compared to reducing the number of parameters.
We showed that a deviation model corresponding to detectable bit-cell faults combined with a bit masking technique can be replaced by a simpler erasure model to speed up the training, and that the use of this regularizer during the training phase allows to further reduce the energy with no impact on accuracy.

Finding the architecture that achieves the best accuracy for a given energy budget still remains a highly open question, considering the very large number of possible solutions.
As such, a more systematic study of the combined impact of pruning, quantizing, factorizing, reducing the number of parameters, tweaking hyperparameters and reducing supply voltage is a very promising direction for future work.

\bibliographystyle{IEEEtran}
\bibliography{IEEEabrv,ref}
\end{document}